\documentclass{vgtc}                          % final (conference style)
\ifpdf%                                % if we use pdflatex
  \pdfoutput=1\relax                   % create PDFs from pdfLaTeX
  \usepackage{graphicx}                % allow us to embed graphics files
  \DeclareGraphicsExtensions{.pdf,.png,.jpg,.jpeg} % for pdflatex we expect .pdf, .png, or .jpg files
\else%                                 % else we use pure latex
  \usepackage{graphicx}                % allow us to embed graphics files
  \DeclareGraphicsExtensions{.eps}     % for pure latex we expect eps files
\fi%

\graphicspath{{figures/}{pictures/}{images/}{./}} % where to search for the images

\usepackage{microtype}                 % use micro-typography (slightly more compact, better to read)
\PassOptionsToPackage{warn}{textcomp}  % to address font issues with \textrightarrow
\usepackage{textcomp}                  % use better special symbols
\usepackage{mathptmx}                  % use matching math font
\usepackage{times}                     % we use Times as the main font
\usepackage{cite}                      % needed to automatically sort the references
\usepackage{tabu}                      % only used for the table example
\usepackage{booktabs}                  % only used for the table example
\usepackage{subfig}
\usepackage{svg}
\renewcommand\footnotemark{}

\usepackage{multirow}

\usepackage{float}

\usepackage{tabularx}

\usepackage{colortbl}
\usepackage{xcolor}
\usepackage{diagbox}  % Include this in your preamble
\usepackage{booktabs} % Optional: for nicer lines
\usepackage{amssymb}
\usepackage{amsmath}
\usepackage{algorithm}
\usepackage{algpseudocode}

\algtext*{EndIf}
\algtext*{EndFor}

\usepackage{booktabs}
\usepackage{pifont}
\usepackage{makecell}
\usepackage{tabularx}
\usepackage{array}

\newcommand{\cmark}{\ding{51}}
\newcommand{\xmark}{\ding{55}}

\newcolumntype{C}[1]{>{\centering\arraybackslash}p{#1}}
\newcolumntype{L}[1]{>{\raggedright\arraybackslash}p{#1}}

\onlineid{xxxx}

\vgtccategory{Research}

\usepackage{amsmath}

\vgtcinsertpkg

\usepackage{hyperref}
\usepackage{cleveref}
\usepackage{mdframed}
\usepackage{makecell}
\usepackage{url}
\usepackage[most]{tcolorbox}
\usepackage{enumitem}

\title{Real-Time Visual Obstruction Detection in Surgical Augmented Reality}

\author{Shih-Chin Yang$^{1*}$%
\and Yanming Xiu$^{2*}$%
\and Hanting Ye$^{2}$%
\and Qi Chen$^{2}$%
\and Elias Rotondo$^{1}$%
\and Maria Gorlatova$^{2}$}
\affiliation{\scriptsize $^1$Department of Computer Science, Duke University \\ $^2$Department of Electrical and Computer Engineering, Duke University\thanks{\parbox{\textwidth}{\noindent *Equal contribution. \\ \{$\mathrm{shih\mbox{-}chin.yang}^{1}, \rm{yanming.xiu}^2, \rm{hanting.ye}^3,\\ \rm{qi.chen2}^4, \rm{eli.rotondo}^5, \rm{maria.gorlatova}^6$\}@duke.edu}}}

\abstract{

Surgical augmented reality (AR) can provide contextual guidance by overlaying virtual annotations, tool cues, and procedural information onto the surgical workspace. However, the virtual content may obstruct task-relevant real-world information, such as surgical instruments, and interfere with users' perception during time-sensitive surgical tasks. In this paper, we investigate visual obstruction detection for surgical AR and present a latency-aware pipeline that combines vision-language model (VLM)-based surgical-object recognition with segmentation-based obstruction reasoning. To reduce inference overhead, the system adopts a cascaded small-to-large VLM architecture with segmentation-guided early exiting and attention-based visual token pruning. The small VLM handles easy frames when its key-object prediction is supported by segmentation consistency, while difficult frames are forwarded to a large VLM with pruned visual tokens. We construct a pseudo-AR surgical obstruction detection benchmark by overlaying virtual content onto surgical-tool images and labeling whether the virtual content obstructs task-relevant instruments. Evaluation results show that the proposed system achieves 87.43\% obstruction detection accuracy with an average end-to-end latency of 479~ms, reducing latency by 62.90\% compared with a cloud large-model baseline. These results demonstrate the feasibility of latency-aware obstruction detection for surgical AR and motivate future work on dynamic surgical videos, multi-object scenes, and clinically grounded AR guidance content.

} % end of abstract

\CCScatlist{
    \CCScatTwelve{Augmented Reality}{Surgical AR}{Obstruction Detection}{Vision-Language Model}{Efficient Inference};
}

\begin{document}


\firstsection{Introduction}

\maketitle

% Key object detection: low acc.

% Obstruction detection: high acc.

% Current: 350 image pairs; accuracy: latency: 1.3 shttps://arxiv.org/pdf/2602.00414

\label{sec:intro}

Augmented Reality (AR) has been increasingly applied to enhance users' perception of the physical world with contextual information, interactive guidance, and real-time feedback. As AR systems become more capable and accessible, they are being explored in a wide range of high-stakes domains beyond entertainment and general-purpose visualization. One important example is surgical AR, where virtual annotations can support real-world procedures and surgical training by highlighting anatomical structures, visualizing tool trajectory paths, presenting procedural instructions, or providing task-specific feedback~\cite{malhotra2023augmented, yoon2018augmented, suresh2023role}. In these settings, AR can improve situational awareness and reduce cognitive burden by integrating useful information directly into the user's visual workspace. 

%\yx{need about 5 citations here. Maybe we can use Sarah's work}

\begin{figure}
    \centering
    \includegraphics[width=1\linewidth]{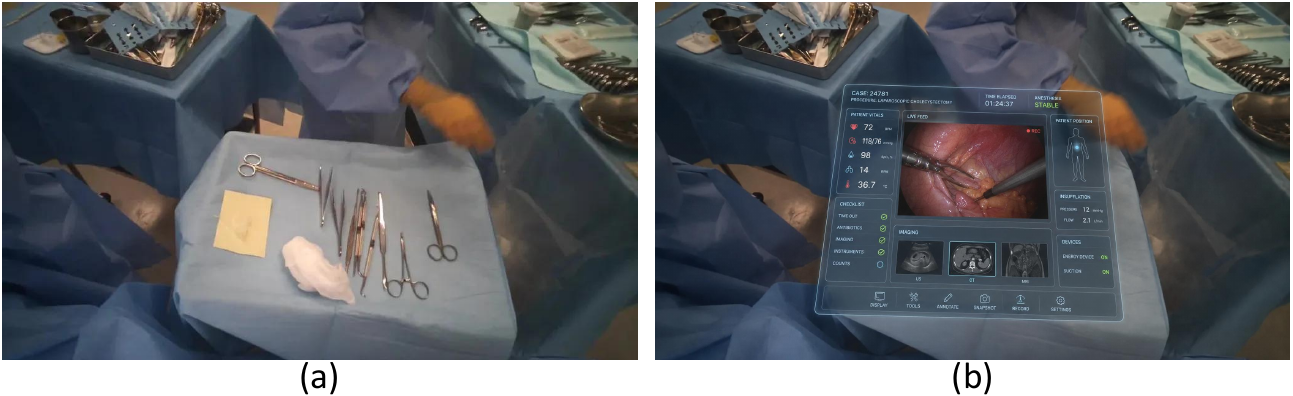}
    \vspace{-0.6cm}
    \caption{Example of visual obstruction in surgical AR. (a) A surgical scene with task-relevant surgical instruments. (b) 
    An AR interface view of a virtual window occluding instrument visibility.
    % An AR view where a virtual interface obstructs the surgical workspace, making the instruments invisible to the user.
    }
    \label{fig:intro}
    \vspace{-0.6cm}
\end{figure}

However, the same virtual content that provides guidance can also interfere with the user's perception of the real-world surgical scene. A particularly important risk is obstruction, where virtual content occludes task-critical real-world information, as shown in~\cref{fig:intro}. 
In surgical AR, elements sensitive to obstructions include anatomical structures, medical instruments, monitor readouts, measurement indicators, or other visual cues needed for safe and effective task execution. 
% In surgical AR, obstructed content may include anatomical structures, surgical instruments, monitor readouts, measurement indicators, or other visual cues needed for safe and effective task execution. 
As surgical tasks require precise visual awareness and rapid responses, even temporary or partial obstructions degrade the user's ability to interpret the scene or manipulate instruments correctly. Detecting such impediments is therefore an important step toward safer and more reliable surgical AR systems.

Prior work has studied obstruction-related problems in AR, including the management of occlusion relationships between virtual content~\cite{obstruction01, obstruction02} and the detection of virtual content that blocks important real-world objects~\cite{obstruction03, obstruction04}. Recent vision-language model (VLM)-based approaches have shown that VLMs can identify task-relevant objects and task-detrimental AR content~\cite{viddar, xiu2025detecting, xiu2026benchmarking}. However, most existing systems are designed for general AR scenarios rather than surgical environments. 
% Surgical AR introduces additional constraints: the system must operate with low latency due to the safety-critical nature of the task. Besides, it must recognize domain-specific objects such as operative tools and anatomical structures, which general-purpose benchmarks may not reliably evaluate.
Surgical AR introduces additional constraints: the system must operate with low latency due to the safety-critical nature of the task, and it must recognize domain-specific objects such as operative tools and anatomical structures.
% ; and it must account for privacy and deployment restrictions that make commercial cloud-based models less desirable.

In this work, we investigate efficient obstruction detection specialized for surgical AR. 
We propose a cascaded small-to-large VLM pipeline that combines segmentation-guided early exiting with attention-based visual token pruning. 
The system uses a lightweight model to handle easy frames when its surgical-tool prediction is supported by segmentation consistency, and falls back to a larger model with pruned visual tokens for more challenging frames. 
This design reduces unnecessary large-model computation while preserving stronger recognition capability for difficult surgical frames. 
This paper makes the following contributions:

\begin{itemize}

    \vspace{-0.2cm}

    \item We construct a pseudo-AR surgical obstruction detection benchmark from a surgical instrument dataset by overlaying virtual content onto surgical-tool images and labeling whether the virtual content obstructs task-relevant instruments. 
    
    \vspace{-0.2cm} 

    \item We propose a cascaded small-to-large VLM pipeline for surgical AR obstruction detection, using segmentation-guided early exiting and attention-based visual token pruning to reduce inference overhead while preserving detection reliability. 

    \vspace{-0.2cm}
    
    \item We evaluate the proposed pipeline and multiple baselines on the constructed benchmark, analyzing detection quality, latency, token-pruning behavior, and early-exit rate. The system achieves 87.43\% obstruction detection accuracy with an average end-to-end latency of 479~ms, reducing latency by 62.90\% compared with the cloud large-model baseline.

    \vspace{-0.2cm}
\end{itemize}

The remainder of this paper is organized as follows: \cref{sec:related work} reviews prior work on surgical AR, AR obstruction detection, and efficient VLM inference. \cref{sec:method} presents our latency-efficient surgical AR obstruction detection pipeline. \cref{sec:evaluation} then describes the experimental setup and evaluates the proposed pipeline against multiple baselines for detection quality, latency, and efficiency.
We discuss the implications and limitations of this pilot-level study in~\cref{sec:future}, and conclude the paper in~\cref{sec:conclusion}.
\section{Related Work}
\label{sec:related work}

\subsection{Surgical AR}

The surgical community leverages AR and head-mounted displays to enhance intraoperative care and medical training~\cite{malhotra2023augmented, yoon2018augmented, suresh2023role, checcucci_metaverse_2026}. By overlaying patient-specific 3D anatomy, medical imaging, and vital signs directly within the surgeon’s line of sight, surgical AR delivers real-time spatial guidance while minimizing cognitive load and gaze redirection. Harnessing these capabilities, surgeons and healthcare institutions are increasingly adopting AR systems into operating room workflows. For example, Augmedics’ xvision Spine System~\cite{augmedics2019xvision} provides an intuitive 3D AR navigation platform for spinal surgery, allowing surgeons to visualize internal structures without looking away from the surgical field. Meanwhile, the OnPoint AR Spine System~\cite{onpoint2023arspine} emphasizes an open architecture, offering third-party implant and instrument compatibility to reduce the overall learning curve. Beyond systems reliant on custom hardware, platforms developed by Medivis and Novarad adopt software-first architectures that leverage commercial off-the-shelf headsets. Medivis’ SurgicalAR~\cite{medivis2019surgicalar} and Novarad’s VisAR~\cite{novarad2022visar} cater to multiple surgical disciplines, spanning neurosurgical and orthopedic procedures. SurgicalAR prioritizes high-resolution 3D soft-tissue rendering for complex anatomy such as brain tumors and blood vessels~\cite{KAHN2025107,GURSES2025108820}. In contrast, VisAR focuses on rigid anatomical tracking and trajectory guidance, enabling surgeons to compare instrument depth directly against overlaid radiology slices~\cite{felix_augmented_2022,evans_improved_2022}. Beyond intraoperative tools, prospective AR systems also support surgical training. Examples include guiding catheter placement in neurosurgery~\cite{eom2025neurolens}, targeting for laser indirect ophthalmoscope retinal therapy~\cite{EOM2026100353}, and reference visualization during wound suturing~\cite{nagayo_augmented_2022,lovett_optimizing_2024}.

% While modern surgical AR aims to optimize operational performance and patient safety, current deployments often encounter perceptual challenges due to visual clutter, improper depth cues, and cumulative tracking errors. Addressing these limitations, our work targets virtual content obstructions within medical environments. In safety-critical surgical workflows, dynamic occlusion management is imperative to preserve the surgeon’s situational awareness and eliminate visual distractions.

While surgical AR aims to improve safety and performance, issues such as visual clutter, improper depth cues, and tracking errors persist. To maintain situational awareness and eliminate distractions in safety-critical surgery environments, our work focuses on dynamic occlusion management to address virtual content obstructions.

\subsection{AR Obstruction Detection}

% Obstructions are established core challenges in AR.
Obstruction is a well-established challenge in AR. Prior work has studied AR obstruction from rendering, interface, and security perspectives. 
Shah et al.~\cite{obstruction01} reviewed occlusion problems in AR, emphasizing the importance of handling visibility relationships between real and virtual objects. 
Davari et al.~\cite{obstruction02} investigated occlusion management techniques for AR interfaces, where virtual content may interfere with the user's awareness of the physical environment. 
% Satkowski et al.~\cite{obstructionismar1} explored ceiling and floor regions as alternative placement areas for AR content to reduce interference with the user's primary field of view. 
From a security and safety perspective, Lebeck et al.~\cite{obstruction03} proposed AR output policies to constrain virtual content based on its relationship to real-world objects, including preventing virtual objects from blocking important physical content. 
Cheng et al.~\cite{obstruction04} studied user reactions and mental models toward perceptual manipulation attacks in mixed reality, showing that misleading virtual output can affect user understanding and behavior. 
More recently, Xiu et al.~\cite{viddar} proposed ViDDAR, a VLM-based system for detecting virtual content that obstructs task-relevant real-world objects.

However, these prior efforts primarily target general AR settings, interface-level content management, or broad AR output safety. Surgical AR introduces additional constraints: the system must recognize domain-specific instruments, reason about obstruction in visually complex surgical scenes, and respond with low latency for time-sensitive tasks. Our work addresses this gap by combining VLM-based surgical-object recognition, segmentation-based obstruction reasoning, and efficient cascaded VLM inference to detect whether virtual content obstructs task-relevant surgical tools.

\subsection{Efficient VLM Inference}

VLMs provide strong multimodal scene understanding solutions, but their high inference cost and latency make efficient inference important for interactive AR systems. Recent work has explored multiple directions for reducing the inference cost of VLMs. 
% Wang et al.~\cite{wang2023efficientvlm} proposed EfficientVLM, which combines knowledge distillation and modal-adaptive pruning to obtain smaller and faster vision-language models. Other methods focus on reducing redundant visual-token computation during inference. 
Chen et al.~\cite{chen2024fastv} proposed FastV, a plug-and-play acceleration method that prunes redundant visual tokens in large VLMs. 
Shang et al.~\cite{shang2025llavaprumerge} presented LLaVA-PruMerge, which adaptively reduces visual tokens by pruning less informative tokens and merging visual information to preserve performance. 
In parallel, early-exit methods reduce computation by allowing models to stop at lower-cost exits when predictions are sufficiently reliable. For example, Bajpai et al.~\cite{bajpai2025free} introduced FREE, an early-exit framework for VLMs that supports input-adaptive inference. 
More closely related to our design, Zhao et al.~\cite{zhao2025astitch} proposed SGL, which uses attention maps from a small VLM to guide visual token pruning in a large VLM, dynamically invoking the large model only when needed.

Our work draws on these efficiency-oriented ideas to build a latency-aware obstruction detection pipeline for surgical AR. Instead of treating efficient VLM inference as a standalone optimization problem, we integrate cascading VLM coordination with the structure of the obstruction detection task. The small VLM first predicts task-relevant surgical objects, and segmentation consistency is used to decide whether the system can exit early. When the large VLM is needed, attention-based visual token pruning reduces the cost of large-model inference. This task-aware combination allows the system to reduce unnecessary computation while retaining stronger recognition capability for challenging surgical frames.
\section{Latency-Efficient Surgical AR Obstruction Detection}

\label{sec:method}

\begin{figure}
    \centering
    \includegraphics[width=0.95\linewidth]{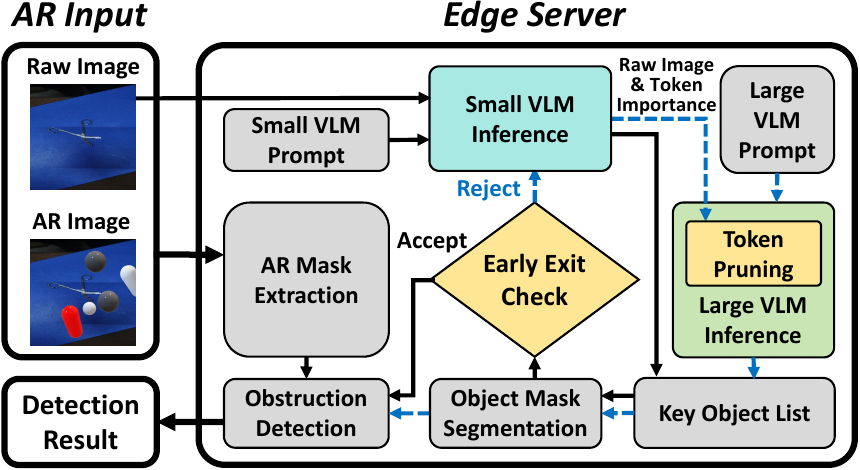}
    \vspace{-0.2cm}
    \caption{Overview of the proposed latency-efficient surgical AR obstruction detection system. Black arrows indicate the default small-model path, while blue arrows indicate the fallback path triggered when the early-exit check is rejected.}
    \label{fig:diagram}
    \vspace{-0.6cm}
\end{figure}

\subsection{System Overview}

Building on recent VLM-based AR obstruction detection approaches~\cite{viddar}, we design a latency-aware pipeline for surgical AR that combines surgical-object recognition, mask-based obstruction reasoning, and a cascaded small-to-large model architecture, as shown in ~\cref{fig:diagram}. The system takes two synchronized inputs from the AR device: a raw surgical image that captures the physical surgical scene, and an AR image that contains the same scene composited with virtual AR content. The goal is to determine whether the virtual content obstructs task-relevant surgical objects in the user's view.

The edge server first processes the raw surgical image with a VLM to identify task-relevant surgical objects, producing a key-object list. 
This list is then used as text prompts for an open-set segmentation model, which takes the current image and object names as input and generates masks for the specified surgical instruments or tools.
In parallel, the system extracts an AR mask by comparing the raw surgical image and the AR image. The AR mask identifies the image regions occupied by virtual content. Finally, the system compares the object masks with the AR mask to determine whether the virtual content overlaps with and obstructs any key object. The resulting obstruction decision is returned to the AR device.

To meet the latency requirements of surgical AR, we adopt a cascaded small-to-large VLM architecture on the edge server. The small VLM provides fast responses and initial key-object predictions, while the large VLM provides stronger visual reasoning capability for more challenging surgical frames. Because the two models differ in both capability and role within the pipeline, we design model-specific prompts for them. The small-VLM prompt encourages concise and precise surgical-object names, which facilitate downstream segmentation and early-exit checking. The large-VLM prompt allows broader object descriptions, enabling more robust recognition when the small VLM prediction may be incomplete or ambiguous. When the small VLM's predictions are sufficiently reliable, the system can make an obstruction decision without invoking the large VLM. Otherwise, the system falls back to the large VLM. This cascaded design reduces unnecessary large-model computation while preserving the ability to handle difficult surgical frames.

Both the small and large VLMs are deployed on the edge server rather than through commercial cloud APIs. This deployment choice reduces communication latency compared with remote cloud inference, which is important for time-sensitive surgical AR applications. 
% Meanwhile, it keeps surgical images within a controlled local computing environment, reducing privacy risks associated with transmitting sensitive medical visual data to external services.

{
\setlength{\textfloatsep}{6pt plus 2pt minus 2pt}
\setlength{\floatsep}{6pt plus 2pt minus 2pt}
\setlength{\intextsep}{6pt plus 2pt minus 2pt}
\begin{algorithm}[!t]
\caption{Segmentation-Guided Early-Exit Validation}
\label{alg:early_exit}
\footnotesize
\setlength{\baselineskip}{0.9\baselineskip}
\begin{algorithmic}[1]
\Require Segmentation masks $\mathcal{M}$, key objects $\mathcal{K}$,
and IoU threshold $\tau$
\Ensure Early-exit decision

\If{$\mathcal{K}=\emptyset$}
    \If{all masks in $\mathcal{M}$ are empty}
        \State \Return \textbf{true}
    \Else
        \State \Return \textbf{false}
    \EndIf
\EndIf

\Repeat
    \State Find $M_i, M_j \in \mathcal{M}$ such that
    $i \neq j$ and $\operatorname{IoU}(M_i,M_j)>\tau$

    \If{such a pair exists}
        \State $M_i \gets M_i \cup M_j$
        \State $\mathcal{M} \gets \mathcal{M} \setminus \{M_j\}$
    \EndIf
\Until{no such pair exists}

\If{$|\mathcal{M}| \neq |\mathcal{K}|$}
    \State \Return \textbf{false}
\EndIf

\State \Return \textbf{true}
\end{algorithmic}

\end{algorithm}
}

\subsection{Implementation of Core Modules}

The cascaded VLM architecture described above relies on two core modules to improve inference efficiency while preserving the ability to handle challenging surgical frames. The first module determines whether the small VLM's prediction is sufficiently reliable for an early exit, and the second module reduces the visual-token processing cost when the large VLM is invoked.

\noindent{\textbf{Segmentation-guided early exiting.}} The early-exiting module determines whether the system can make an obstruction decision using only the small VLM. Given a raw surgical image, the small VLM first predicts a list of task-relevant surgical objects. The predicted object names are then used to guide segmentation, producing a set of candidate object masks. The segmentation-guided early-exit validation procedure is summarized in \Cref{alg:early_exit}. Because segmentation may produce multiple masks for the same object, we perform mask-level clustering based on pairwise mean Intersection-over-Union (mIoU). Masks with high mutual overlap are grouped together and treated as one detected object, reducing duplicate detections caused by repeated or fragmented segmentation outputs.

After mask clustering, the system compares the number of predicted key objects with the number of clustered object masks. If they match, the small VLM's prediction is treated as segmentation-consistent, and the system exits early without invoking the large VLM. The clustered object masks are then compared with the AR mask to produce the final obstruction decision. If the two counts do not match, the frame is treated as ambiguous or difficult, and the system rejects the early exit and forwards the frame to the large VLM for stronger surgical-object recognition.

\noindent{\textbf{Attention-based visual token pruning.}} When early exiting is rejected, the large VLM is invoked to improve recognition reliability on difficult frames. To reduce the cost of this large-model path, we adopt small-VLM-guided visual token pruning, following the observation that attention maps from a small VLM can provide effective guidance for pruning visual tokens in a larger VLM~\cite{zhao2025astitch}. During small-VLM inference, we derive token-level importance scores from the small VLM's attention maps. These scores form an importance vector with the same length as the visual-token sequence, where each score indicates the estimated relevance of the corresponding visual-token position to the current prompt and prediction.

The importance scores are used to rank visual-token positions. Given a predefined visual-token retention ratio, the system prunes lower-ranked tokens during large-VLM inference. Importantly, the visual-token embeddings produced by the small VLM are not directly transferred to the large VLM. Instead, the large VLM still encodes the input image using its own visual encoder, and pruning is applied to the large VLM's own visual tokens according to the importance ranking derived from the small VLM. This avoids cross-model token incompatibility while using the small VLM as a guide for reducing large-model computation.

We use decode-stage attention as the attention source for token-pruning guidance. This choice aligns with our early-exit pipeline because the small VLM already performs generation to produce the key-object list, allowing decode-stage attention to be collected during the initial small-model pass. By contrast, extracting prefill-stage attention for all frames would add overhead even for frames that exit early, while extracting it only after early-exit rejection requires an additional partial forward pass. The decode-stage configuration therefore better matches our latency-oriented system design. The overall procedure is summarized in \Cref{alg:token-pruning}.

{
\setlength{\textfloatsep}{6pt plus 2pt minus 2pt}
\setlength{\floatsep}{6pt plus 2pt minus 2pt}
\setlength{\intextsep}{6pt plus 2pt minus 2pt}
\begin{algorithm}[t]
\caption{Small-VLM-Guided Visual Token Pruning}
\label{alg:token-pruning}
\footnotesize
\setlength{\baselineskip}{0.9\baselineskip}
\begin{algorithmic}[1]
\Require Image $I$, small VLM $\mathcal{M}_S$, large VLM $\mathcal{M}_L$,
small-model prompt $q_S$, large-model prompt $q_L$, retention ratio $\rho$
\Ensure Large-VLM prediction $\hat{y}_L$

\Statex \textbf{// Phase 1: Extract visual token importance from small VLM}
\State Run $\mathcal{M}_S$ on $(I,q_S)$ to generate $\hat{y}_S$ while
collecting decode-stage visual attention
\State $N_G \gets$ number of generated tokens in $\hat{y}_S$
\State $N_V \gets$ number of visual tokens processed by $\mathcal{M}_S$
\State $L_S,H_S \gets$ numbers of layers and attention heads in $\mathcal{M}_S$
\State $\mathbf{A}_{t,\ell,h} \in \mathbb{R}^{N_V}$: visual-token attention
at decoding step $t$, layer $\ell$, and head $h$
\State Aggregate the collected attention into importance vector $\mathbf{s}$:
\[
\mathbf{s}
\gets
\sum_{t=1}^{N_G}
\sum_{\ell=1}^{L_S}
\sum_{h=1}^{H_S}
\mathbf{A}_{t,\ell,h}
\]

\Statex \textbf{// Phase 2: Prune large-VLM visual tokens by importance}
\State $k \gets \max(1,\lfloor \rho N_V \rfloor)$
\State $\mathcal{K} \gets$ indices of the top-$k$ entries of $\mathbf{s}$,
ordered by their original visual-token positions
\State $\mathbf{Z}_L \gets$ encode $I$ using the visual encoder of $\mathcal{M}_L$
\State $\mathbf{Z}_L' \gets$ large-VLM visual tokens at positions $\mathcal{K}$
% \State $\hat{y}_L \gets \mathcal{M}_L.\operatorname{Generate}(q_L,\mathbf{Z}_L')$
\State Run $\mathcal{M}_L$ using $q_L$ and $\mathbf{Z}_L'$ to obtain $\hat{y}_L$
\State \Return $\hat{y}_L$
\end{algorithmic}
\end{algorithm}
}

\section{System Evaluation}

\label{sec:evaluation}

With the proposed detection system, we further evaluate it on a pseudo-AR surgical-tool dataset. The evaluation is designed to assess both detection performance and inference efficiency. In this section, we first describe how the dataset is constructed by adding virtual content to surgical-tool images and assigning obstruction labels. We then present the experimental setup, including the VLMs, segmentation model, deployment configurations, and evaluation metrics. Finally, we compare the proposed cascaded VLM pipeline with cloud-based, small-only, and large-only baselines, and analyze its impact on obstruction detection accuracy and latency.

\begin{table*}[t]
\centering
\caption{Comparison of detection accuracy and latency across model configurations.}
\vspace{-0.3cm}
\label{tab:evaluation_quality}
% \small
\footnotesize
\renewcommand{\arraystretch}{0.95}
\begin{tabularx}{\textwidth}{
C{0.04\textwidth}
C{0.10\textwidth}
C{0.04\textwidth}
C{0.06\textwidth}
C{0.06\textwidth}
C{0.13\textwidth}
C{0.13\textwidth}
C{0.12\textwidth}
C{0.12\textwidth}
}
\toprule
\textbf{Model host} &
\textbf{Model \quad \quad configuration} &
\textbf{Early exit} &
\textbf{Token pruning} &
\textbf{Attention source} &
\textbf{Key object detection accuracy} &
% \textbf{Key object detection exact match acc.} &
% \textbf{Obstruction detection acc. w/ key object match} &
\textbf{Obstruction detection accuracy} &
\textbf{Key object detection latency} &
\textbf{End-to-end detection latency} \\
\midrule
Cloud & Large-only & N/A & N/A & N/A
& 0.7486 & 0.8114 & 1119 ms & 1292 ms \\

Edge & Small-only & \xmark & \xmark & N/A
& 0.7229 & 0.7886 & \textbf{156 ms} & \textbf{304 ms} \\

Edge & Large-only & \xmark & \xmark & N/A
& \textbf{0.9257} & 0.7971 & 269 ms & 423 ms \\

Edge & Small + large & \cmark & \xmark & N/A
& 0.7514 & \textbf{0.8743} & 290 ms & 488 ms \\

Edge & Small + large & \xmark & \cmark & Decode
& 0.7000 & 0.7629 & 504 ms & 657 ms \\

% Edge & Small + large & \cmark & \cmark & Prefill
% & 0.7514 & 0.8743 & 0.2875 s & 0.4842 s \\

\textbf{Edge} & \textbf{Small + large} & \cmark & \cmark & Decode
& 0.7514 & \textbf{0.8743} & 279 ms & 479 ms \\
\bottomrule
\end{tabularx}
\label{tab:eval}
\vspace{-0.4cm}
\end{table*}

\subsection{Pseudo-AR Surgical Obstruction Dataset}

% The dataset used in this study was constructed from the surgical tools dataset collected by Reddy et al.~\cite{reddy2025surgicaltools}. We selected 350 single-object images from seven surgical tool categories and added AR-like virtual content to each image. The virtual elements were randomly placed to introduce variation in their spatial relationship with the surgical tools. For each augmented image, obstruction labels were manually annotated based on the overlap between the virtual content and the surgical tool, and an image was labeled as obstructed when the overlapping area could affect human identification of the tool. 

% The original dataset contains 6,000 images across nine surgical instrument categories, such as forceps, hemostats, scalpels, and Mayo scissors. The images were captured on surgical drapes under different lighting conditions and viewing angles to reflect realistic surgical scenarios. These characteristics make the dataset suitable for evaluating surgical tool recognition in visually diverse and clinically relevant environments. By adding virtual content, our dataset further supports the evaluation of AR-assisted perception tasks where virtual elements may interact with or obstruct real surgical instruments. Some data samples from the dataset are shown in~\cref{fig:dataset}.

The dataset used in this study was constructed from the Surgical Tools dataset collected by Reddy~\cite{reddy2025surgicaltools}. The original dataset contains 6,000 images across nine surgical instrument categories, including forceps, hemostats, scalpels, and Mayo scissors. The images were captured on surgical drapes under different lighting conditions and viewing angles, making the dataset suitable for evaluating surgical tool recognition in visually diverse surgical scenes.

\begin{figure}
    \centering
    \includegraphics[width=1\linewidth]{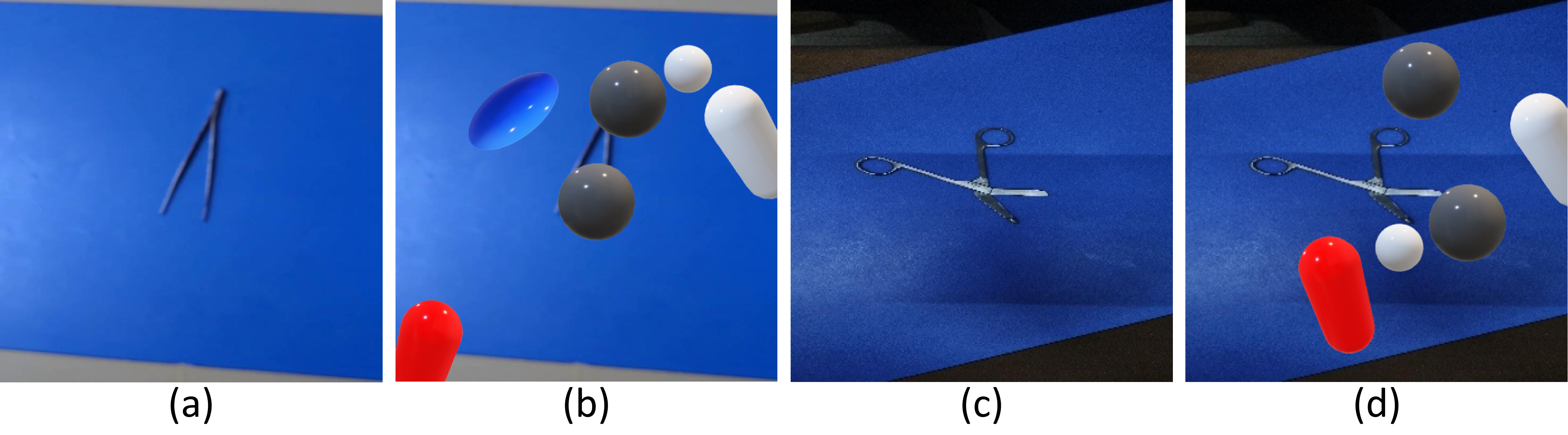}
    \vspace{-0.6cm}
    \caption{Examples from the surgical obstruction detection dataset. (a) A raw surgical image containing forceps, and (b) its AR-composited version; this pair is labeled as obstructed. (c) A raw surgical image containing episiotomy scissors, and (d) its AR-composited version; this pair is labeled as not obstructed.}
    \vspace{-0.6cm}
    \label{fig:dataset}
\end{figure}

From this dataset, we selected 350 single-object images from seven surgical tool categories to construct a pseudo-AR surgical obstruction benchmark. For each selected image, we added AR-like virtual content with randomized placement to introduce variation in its spatial relationship with the surgical tool. Obstruction labels were manually annotated based on the overlap between the virtual content and the target tool. An image was labeled as obstructed when the virtual content covered a region of the tool that could affect human identification; otherwise, it was labeled as not obstructed. By adding virtual content to real surgical-tool images, the constructed benchmark supports evaluation of AR-assisted perception tasks where virtual elements may interact with or obstruct real surgical instruments. Example image pairs from the dataset are shown in~\cref{fig:dataset}.

\subsection{Evaluation Setup and Metrics}

The system was evaluated on a workstation with two NVIDIA RTX A6000 GPUs. The evaluation used two VLMs, Qwen3-VL-2B as the small model and Qwen3-VL-8B as the large model~\cite{qwen3vl2025}, along with SAM3~\cite{carion2025sam3} for text-guided segmentation. 
We selected general-purpose models because our task requires open-vocabulary recognition and a matched small-large model pair; existing surgical segmentation models are commonly optimized for predefined instrument categories, while surgical VLMs are often specialized for bounded tasks such as phase, action, and tool recognition, as well as surgical visual question answering~\cite{yue2024surgicalsam,wang2025surgical}.
GPT-5.4-2026-03-05~\cite{openai2026gpt54} running in the cloud was used as the baseline.

To compare different deployment strategies, we evaluated several system configurations. The large-only cloud setting sends all images to the cloud model and serves as the reference for cloud-based performance. The large-only edge setting uses Qwen3-VL-8B locally to examine whether the cloud model can be replaced while reducing latency. The small-only edge setting uses Qwen3-VL-2B for all cases to assess whether a small-to-large coordination mechanism is necessary. In addition, we evaluated early exiting and token pruning strategies to analyze their impact on accuracy and latency.

The systems were evaluated using accuracy and latency metrics for both key object detection and obstruction detection. Key object detection accuracy measures the VLM's ability to identify the target surgical tool. To compute this metric, we manually defined an acceptable-name list for each surgical-tool class, including semantically equivalent names and descriptions. A VLM prediction is counted as correct if its generated object list contains at least one name from the corresponding list, rather than requiring an exact string match with the dataset category. Obstruction detection accuracy reflects both the performance of SAM3 and the precision of the VLM output used to guide segmentation. Accuracy is reported as a percentage, and latency is measured in seconds.

\subsection{Experimental Results and Analysis}

Based on the dataset and experimental setup described above, we evaluate the proposed cascaded small-to-large VLM pipeline and compare it with several baselines. The evaluation focuses on two aspects: whether the system can accurately detect obstruction in pseudo-AR surgical images, and whether it can reduce inference latency for interactive surgical AR use. \Cref{tab:eval} summarizes the accuracy and latency results across all evaluated configurations. We provide a detailed analysis of each configuration below.

\noindent{\textbf{Proposed system configuration.}} The proposed system configuration, which combines segmentation-guided early exiting and decode-stage token pruning, achieves an obstruction detection accuracy of 87.43\% with an average end-to-end latency of 479~ms. This result suggests that the proposed tiered small-to-large VLM pipeline can provide feasible near-real-time obstruction detection for surgical AR while maintaining strong detection performance.

\noindent{\textbf{Cloud large-only baseline.}} The cloud-based large-model baseline has the highest latency among all evaluated configurations. It requires 1119~ms for key-object detection and 1292~ms for end-to-end obstruction detection. In contrast, the proposed edge-based pipeline reduces end-to-end latency by 62.9\% compared with the cloud baseline, while also improving obstruction detection accuracy from 81.14\% to 87.43\%. This result indicates that relying on a cloud VLM does not necessarily provide the best accuracy, while it introduces substantial communication and inference latency. 
% For time-sensitive surgical AR applications, edge deployment is therefore important not only for privacy, but also for practical responsiveness.
For time-sensitive surgical AR applications, these results highlight the latency advantage of edge deployment.

\noindent{\textbf{Edge small-only baseline.}} The edge small-only configuration yields the lowest latency, with 156~ms key-object detection latency and 304~ms end-to-end latency. However, its detection performance is weaker, achieving 72.29\% key-object detection accuracy and 78.86\% obstruction detection accuracy. Compared with the proposed configuration, the small-only baseline is 175~ms faster in end-to-end latency, but its obstruction detection accuracy is lower by 8.57\%. This result shows that the small-only VLM is efficient but unreliable for surgical-tool recognition and downstream obstruction detection.

% \noindent{\textbf{Edge large-only baseline.}} The edge large-only configuration achieves the highest key-object detection accuracy, reaching 92.57\%, which is 17.71\% higher than the cloud large-model baseline. A closer inspection suggests that this difference is partly due to output granularity: in visually challenging cases, the edge large model often produces broader labels such as ``surgical tool,'' which are accepted under our class-level matching criterion, while the cloud model sometimes attempts more fine-grained but incorrect instrument names. However, this stronger key-object detection does not directly translate into stronger obstruction detection. The edge large-only configuration achieves only 79.71\% obstruction detection accuracy, which is 7.72\% lower than the proposed system. This gap shows that key-object detection and obstruction detection are related but not identical tasks. Obstruction detection also depends on whether the VLM output is suitable for text-guided segmentation and subsequent mask-overlap reasoning.

\noindent{\textbf{Edge large-only baseline.}} The edge large-only configuration achieves the highest key-object detection accuracy, reaching 92.57\%. However, its obstruction detection accuracy is only 79.71\%, which is lower than the proposed system by 7.72\%. This gap shows that key-object detection and obstruction detection are related but not identical tasks. Obstruction detection depends not only on whether the VLM identifies the correct tool category, but also on whether its output is suitable for text-guided segmentation and subsequent mask-overlap reasoning. A model that performs well at object recognition may still produce outputs that are overly specific, ambiguous, or less aligned with the segmentation model, leading to weaker final obstruction detection performance.

\noindent{\textbf{Token-pruning-only baseline.}} The token-pruning-only configuration performs worst in both accuracy metrics, with 70.00\% key-object detection accuracy and 76.29\% obstruction detection accuracy. Its end-to-end latency is also 657~ms, which is slower than both the large-only edge baseline and the proposed configuration. This result suggests that token pruning alone is not sufficient when the system always relies on the large-model path without using the small model as a front filter. One possible reason is that different model sizes may be better suited for different samples: some frames can be handled reliably by the small VLM, while others require the large VLM. 
Without early exiting, the system cannot adapt dynamically, increasing the risk that pruning removes visual details critical for difficult cases.
% Without early exiting, the system loses this adaptive behavior, and pruning may further remove visual information that is useful for difficult cases.

\noindent{\textbf{Impact of token pruning.}} Adding token pruning to early exiting preserves detection performance while reducing latency. The early-exit-only configuration and the proposed configuration achieve the same key-object detection accuracy of 75.14\% and obstruction detection accuracy of 87.43\%. However, adding token pruning reduces key-object detection latency from 290~ms to 279~ms, a 3.6\% reduction, and reduces end-to-end latency from 488~ms to 479~ms. Although the latency improvement is modest, it demonstrates that small-VLM-guided token pruning can reduce large-model inference overhead without degrading detection accuracy in this pipeline.

\begin{figure}
    \centering
    \includegraphics[width=1\linewidth]{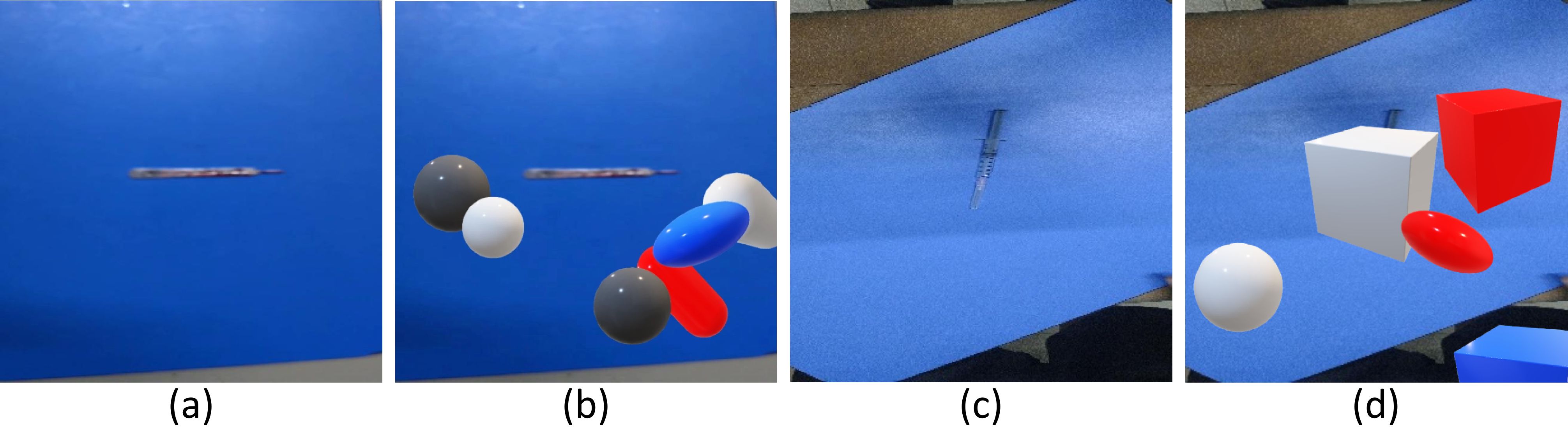}
    \vspace{-0.65cm}
    \caption{Representative obstruction detection failure cases. (a,b) A scalpel example where failure is caused by incorrect key-object detection. (c,d) A syringe example where the key object is correctly identified, but segmentation fails to produce a valid object mask.}
    \vspace{-0.6cm}
    \label{fig:failure}
\end{figure}

\subsection{Failure Analysis}

To better understand the obstruction detection failures, we examine representative failure cases, as shown in~\cref{fig:failure}. Since the proposed pipeline detects obstruction through a sequence of key-object recognition, object-mask segmentation, and mask-overlap reasoning, failure cases can originate from different stages of the pipeline. In our inspection, we observe two major sources of obstruction detection failure: key-object detection failure and segmentation failure.

\noindent{\textbf{Key-object detection failure.}} \Cref{fig:failure}(a,b) shows a failure case involving a scalpel. In this example, the VLM incorrectly predicts the target object as a thermometer, resulting in a false positive obstruction detection. This type of error commonly occurs when the input image has low visual clarity, weak object boundaries, or motion blur. Since the scalpel is thin and visually subtle, even a small amount of blur can make it difficult for the VLM to distinguish the tool from the background. Once the key object is incorrectly identified, the downstream segmentation module receives an incorrect text prompt, which prevents the system from localizing the correct surgical instrument and leads to an incorrect obstruction decision.

\noindent{\textbf{Segmentation failure.}} \Cref{fig:failure}(c,d) shows a failure case in which the VLM correctly identifies the key object as a syringe, but SAM3 fails to detect the object and returns no valid mask. This failure may be caused by low image clarity, the small and elongated shape of the syringe, or the use of relatively specialized object names as text prompts. Since SAM3 is a general-purpose segmentation model, its training coverage for fine-grained surgical instruments may be limited compared with common everyday objects. As a result, even when the VLM provides a correct key-object label, the segmentation stage may fail to localize the instrument, causing the obstruction detection pipeline to miss the obstruction.

These failure cases indicate that obstruction detection depends on both semantic recognition and mask-level localization. Improving the system therefore requires not only stronger VLM-based surgical-tool recognition, but also segmentation models or prompt strategies that are better adapted to surgical instruments, low-quality surgical images, and visually subtle tool structures.
\section{Discussion and Future Work}

\label{sec:future}

Our evaluation shows that the proposed cascaded small-to-large VLM pipeline can support latency-aware obstruction detection for surgical AR, achieving strong obstruction detection accuracy while substantially reducing end-to-end latency compared with the cloud large-model baseline. At the same time, this pilot study also reveals several limitations that should be addressed before such systems can be evaluated in more realistic surgical AR settings. We structure the discussion around three limitations that naturally motivate our future work: the simplicity of the current task scenario, the limited scale and static nature of the current dataset, and the remaining inference overhead in frame-by-frame VLM processing.

\noindent{\textbf{Realistic surgical AR scenes and visual complexity.}}
% \noindent{\textbf{Task-relevant target recognition in complex surgical AR scenes.}}
The current benchmark contains only basic AR overlays and a single surgical tool in each image, which simplifies key-object recognition, segmentation, and obstruction reasoning. However, in realistic intraoperative settings, visual complexity arises from both the underlying surgical scene and the overlaid AR content. Multiple surgical tools, hands, and anatomical structures will be present simultaneously~\cite{kenngott2014real}, while AR systems additionally display patient-specific information, such as ultrasound, mammography, CT, and MRI data, within or near the surgical field\cite{chopra2024ARsurgery,ma2023ARreview}. Such overlays can improve information accessibility and spatial guidance by reducing the need to repeatedly consult external monitors, but they also introduce additional visual elements into an already complex operative environment \cite{ma2023ARreview}. In practical systems, virtual displays and patient-anchored overlays often appear in close proximity to physical objects, partially overlap with them, or mutually occlude them\cite{vonHaxthausen2023Ultrasound,heining2024PelvicAR,vanIsseldyk2026ARspine}. These conditions make object recognition, segmentation, and obstruction reasoning substantially more difficult than in the simplified setting considered in this work.

 % Surgical AR visualization can generally be divided into non-in-situ and patient-anchored approaches \cite{ma2023ARreview}. Non-in-situ systems present information on virtual displays within the surgeon’s field of view. For example, Van Isseldyk et al. displayed endoscopic surgical information through an AR headset and reported reduced workload and improved ergonomic posture compared with conventional monitors \cite{vanIsseldyk2026ARspine}. In contrast, patient-anchored systems spatially align virtual information with the operative area. Von Haxthausen et al., for instance, visualized real-time in-situ 3D ultrasound for vascular puncture guidance \cite{vonHaxthausen2023Ultrasound}, while Heining et al. overlaid planned screw trajectories and pelvic anatomy directly onto the surgical field \cite{heining2024PelvicAR}.

Extending the system to realistic surgical AR environments is therefore an important direction for future work. This extension requires changes at both the dataset and system levels. At the dataset level, future benchmarks will focus on creating more realistic representations of surgical scenes, particularly in terms of object interactions, spatial relationships, and occlusion patterns. Currently, we are extending our obstruction detection system to gallbladder removal surgery as a clinically relevant use case. At the system level, the detection pipeline will be refined to provide more reliable obstruction assessment and better support the surgeon’s visual awareness in complex surgical scenes. Specifically, the system will jointly model multiple key objects, their corresponding segmentation masks, and their individual obstruction states. We will also continue to follow advances in surgical foundation models and evaluate more general-purpose variants as they become suitable for open-set obstruction detection.

 \begin{figure}
    \centering
    \includegraphics[width=1\linewidth]{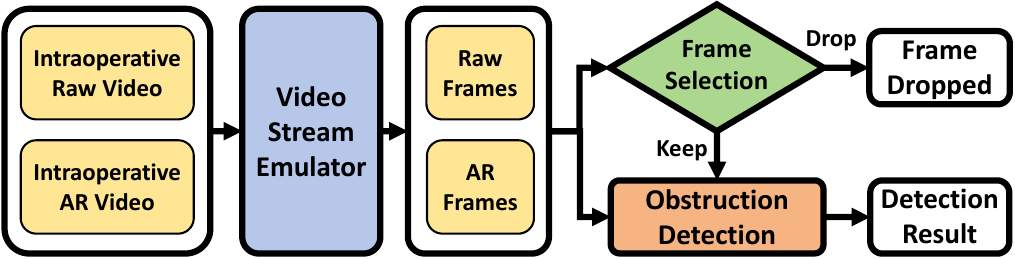}
    \vspace{-0.55cm}
    \caption{Future extension of the surgical AR obstruction detection pipeline. Recorded real-world intraoperative videos and their AR versions are replayed by a video stream emulator to generate paired raw and AR frame streams. A frame-selection module filters out redundant frames to reduce unnecessary inference, while selected frames are forwarded to the obstruction detection module.}
    \vspace{-0.6cm}
    \label{fig:future_ext}
\end{figure}

\noindent{\textbf{Dataset scale and video-based benchmark construction.}} Our current benchmark is a pilot-scale pseudo-AR dataset created by overlaying virtual content onto static images of surgical tools. While this controlled setup supports the evaluation of obstruction detection, the dataset remains limited in size and relies on manually annotated obstruction labels without clinical validation. In addition, its static nature does not reflect the temporal complexity of surgical AR, such as camera and tool motion, viewpoint changes, hand movement, and transient occlusions.

A natural next step is to construct video-based benchmarks for surgical AR obstruction detection. Real surgical AR videos, however, are difficult to collect at scale because of patient privacy, limited operating-room availability, and the need for medical supervision~\cite{silas2015video, quach2023ethical}. AI-based video generation provides one possible way to expand the dataset while maintaining control over virtual content and obstruction conditions. Starting from real surgical images or videos, generative models could synthesize specific surgical moments with controllable AR overlays, camera motion, viewpoint changes, and tool movement~\cite{cho2024surgen, chen2025surgsora}. Compared with manually adding AR overlays to static images, video generation would support more diverse and dynamic evaluation scenarios.

Beyond capturing temporal dynamics, we will also expand the video-based benchmark to cover a broader range of clinically relevant AR guidance. Our current benchmark focuses only on obstructions of surgical instruments, whereas surgical AR can include anatomy labels, highlighted lesions, procedure-specific visual cues, and other overlays~\cite{chan2021integrated, li2024navigate}.
A video-based benchmark could evaluate whether dynamic AR guidance obstructs not only surgical instruments, but also anatomical structures and other task-relevant information. 
% Medical professional supervision will remain essential to ensure that generated content is anatomically plausible, clinically relevant, and appropriate for surgical AR evaluation. 
We will also collaborate with medical professionals to review the generated surgical images and videos to ensure that they are anatomically plausible, clinically relevant, and appropriate for surgical AR evaluation.
By expanding from static pseudo-AR images to generated or recorded video sequences, our future datasets will better capture the visual complexity of surgical workflows while reducing dependence on large-scale real-world surgical recordings.

\noindent{\textbf{Inference efficiency and frame selection.}} Although the tiered small-to-large VLM pipeline reduces unnecessary large-model computation, the current system still invokes VLM-based obstruction detection on every candidate frame. This is not necessary in video-based AR settings. When the user's viewpoint is stable, consecutive frames may contain only minor visual changes, and repeatedly running the full VLM pipeline can introduce unnecessary latency and energy cost. This limitation motivates a frame-selection stage that filters video frames before VLM inference.

In our future work, we will develop a lightweight frame-selection module that uses inexpensive visual cues, such as feature matching~\cite{liu2025liftfeat}, color-histogram comparison~\cite{lubwama2025wifilter}, optical-flow magnitude~\cite{wang2024sea}, or frame-level similarity~\cite{sun2025mdp3, wang2025videotree}, to decide whether a frame should be retained or dropped as redundant. Selected frames will be forwarded to the obstruction detection module, while dropped frames could either reuse previous detection results or be skipped depending on the application requirements. This design would reserve VLM inference for informative frames, reducing computation while preserving responsiveness.

\begin{figure}
    \centering
    \includegraphics[width=1\linewidth]{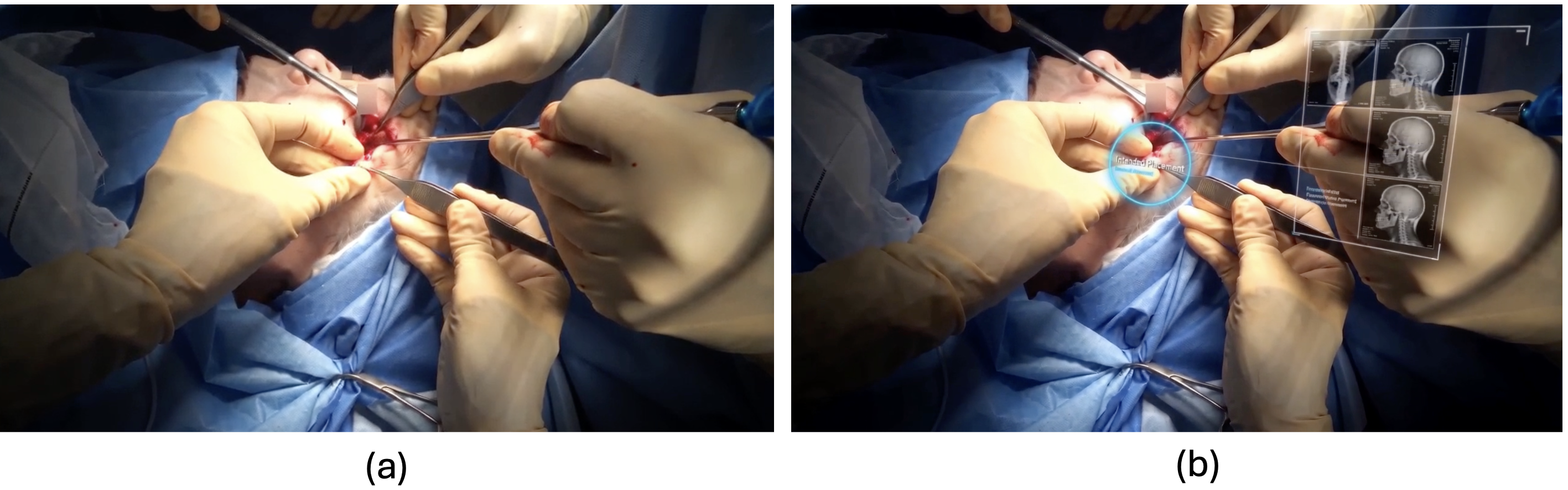}
    \vspace{-0.65cm}
    \caption{Example of generated surgical AR content. (a) A surgical workspace. (b) An AR-enhanced version with virtual medical guidance panels, illustrating the potential of generative methods for creating more complex surgical AR scenarios.}
    \vspace{-0.6cm}
    \label{fig:generated}
\end{figure}

To evaluate such a video-based pipeline, we will replay recorded surgical videos as real-time frame streams. A video stream emulator can decode recorded raw and AR videos and feed paired raw/AR frames into the frame-selection and obstruction-detection modules. A media-processing backend such as Rockchip MPP~\cite{rockchip_mpp} could support this replay process by providing efficient video decoding and frame-stream generation. This setup would allow us to test frame selection under controlled yet dynamic conditions before collecting real-time surgical AR data.

Beyond whole-frame filtering, we will also explore more fine-grained evidence preservation. Classical visual cues operate at the frame-level~\cite{lubwama2025wifilter, chanwut2026tetris, shen2024longvu} and can be too coarse to distinguish benign viewpoint changes from changes that affect task-relevant objects. A tile-level or region-level evidence layer could localize where meaningful changes occur within a frame and aggregate this information over time. Such evidence could then be coupled with our token-pruning strategy, allowing the VLM to focus on regions most likely to contain obstruction. We expect this combination of frame selection, local evidence preservation, and token pruning to further reduce VLM invocations while maintaining detection reliability.
\section{Conclusion}

\label{sec:conclusion}

In this work, we investigate visual obstruction detection for surgical AR, where virtual guidance content may block task-relevant surgical instruments and interfere with user perception. We propose a latency-aware small-to-large VLM pipeline that combines segmentation-guided early exiting with attention-based visual token pruning. Our evaluation on a pseudo-AR surgical obstruction benchmark shows that the proposed system achieves 87.43\% obstruction detection accuracy with an average end-to-end latency of 479~ms, reducing latency by 62.9\% compared with the cloud large-model baseline. The results suggest that efficient VLM-based reasoning can support near-real-time obstruction detection in surgical AR, while our failure analysis shows that remaining errors are mainly caused by key-object recognition failures and text-guided segmentation failures. 
% Building on this study, future work will expand the benchmark to dynamic surgical videos, realistic surgery scenes, and more clinically grounded AR guidance content, while improving frame selection and segmentation strategies for visually subtle surgical instruments.
Building on this study, future work will expand the benchmark to dynamic surgical videos, complex surgical scenes, and more clinically grounded AR guidance content, while improving frame selection and segmentation strategies for task-relevant surgical objects and anatomical structures.

\acknowledgments{
This work was supported in part by NSF grants CSR-2312760, CNS-2112562, and IIS-2231975, NSF CAREER Award IIS-2046072, NSF NAIAD Award 2332744, a CISCO Research Award, a Meta Research Award, Defense Advanced Research Projects Agency Young Faculty Award HR0011-24-1-0001, and the Army Research Laboratory under Cooperative Agreement Number W911NF-23-2-0224. The views and conclusions contained in this document are those of the authors and should not be interpreted as representing the official policies, either expressed or implied, of the Defense Advanced Research Projects Agency, the Army Research Laboratory, or the U.S. Government. This paper has been approved for public release; distribution is unlimited. No official endorsement should be inferred. The U.S.~Government is authorized to reproduce and distribute reprints for Government purposes notwithstanding any copyright notation herein.
}

\bibliographystyle{abbrv-doi}
%\bibliographystyle{abbrv-doi-narrow}
%\bibliographystyle{abbrv-doi-hyperref}
%\bibliographystyle{abbrv-doi-hyperref-narrow}

% \bibliography{template}

\end{document}